\documentclass{bmvc2k}

\usepackage{xcolor}
\usepackage{color}
\usepackage{verbatim} 
\usepackage{enumitem}
\usepackage{multirow} 

\newcommand\XP[1]{\textcolor{black}{#1}}

\newcommand\wch[1]{\textcolor{black}{#1}}

\title{Multi-modal Crowd Counting via Modal Emulation}

\addauthor{Chenhao Wang}{22B903078@stu.hit.edu.cn}{1}
\addauthor{Xiaopeng Hong}{hongxiaopeng@ieee.org}{1}
\addauthor{Zhiheng Ma}{zh.ma@siat.ac.cn}{2}
\addauthor{Yupeng Wei}{23s003027@stu.hit.edu.cn}{1}
\addauthor{Yabin Wang}{iamwangyabin@stu.xjtu.edu.cn}{3}
\addauthor{Xiaopeng Fan}{fxp@hit.edu.cn}{1}

\addinstitution{
 Faculty of Computing, Harbin Institute of Technology, P. R. China
}
\addinstitution{
Shenzhen Institute of Advanced Technology, Chinese Academy of Science, P. R. China
}
\addinstitution{
School of Cyberspace and Engineering, Xi’an Jiaotong University, P. R. China
}
\runninghead{Wang, et al.}{Multi-modal Crowd Counting via Modal Emulation}

\begin{document}

\maketitle

\begin{abstract}
Multi-modal crowd counting is a crucial task that uses multi-modal cues to estimate the number of people in crowded scenes. To overcome the gap between different modalities, we propose a modal emulation-based two-pass multi-modal crowd-counting framework that enables efficient modal emulation, alignment, and fusion. The framework consists of two key components: a \emph{multi-modal inference} pass and a \emph{cross-modal emulation} pass. The former utilizes a hybrid cross-modal attention module to extract global and local information and achieve efficient multi-modal fusion. The latter uses attention prompting to coordinate different modalities and enhance multi-modal alignment. We also introduce a modality alignment module that uses an efficient modal consistency loss to align the outputs of the two passes and bridge the semantic gap between modalities. Extensive experiments on both RGB-Thermal and RGB-Depth counting datasets demonstrate its superior performance compared to previous methods. Code available at https://github.com/Mr-Monday/Multi-modal-Crowd-Counting-via-Modal-Emulation.

\end{abstract}

\section{Introduction}
\label{sec:intro}

Crowd counting is an essential research topic in the field of machine perception. The goal of this task is to accurately estimate the number of people in an image. Over the past decade, crowd counting has been widely used in various fields~\cite{xiong2017spatiotemporal,liu2020dynamic}. Existing crowd counting methods mainly focus on the visual features of RGB images~\cite{liu2018crowd,liu2018attentive,ma2019bayesian,liu2020efficient,ma2020learning,ma2021towards,lin2022boosting} may have limitations when confronted with intricate environments like occlusions and shadows.

Recently, multi-modal crowd counting has gained increasing attention for its ability to address the limitations of using only the visible modality. Fusing thermal or depth images with RGB images can significantly improve counting performance, especially in challenging scenes with low light and occlusion. Most of the previous multi-modal crowd counting approaches \cite{lian2019density,zhou2023mc} are based on convolutional structures for multi-modal fusion, and show that integrating thermal or depth images with RGB images improves crowd counting performance. Nevertheless, simple fusion methods \cite{tang2022tafnet,liu2021cross} are limited in their ability to fully capture the complementarity between modalities. In recent studies, the effectiveness of the Transformer model in multi-modal tasks has been demonstrated \cite{lu2019vilbert,tan2019lxmert}. For example, Wu et al.~\cite{wu2022multimodal} proposes a Mutual Attention Transformer (MAT) method that utilizes a cross-attention mechanism to capture the complementarity of different modalities for robust crowd counting. However, existing methods have not fully explored the issue of modality alignment, which hinders the ability to effectively model multi-modal interaction.

This paper tackles the multi-modal counting problem from a new perspective, arguing that a superior multi-modal feature encoder should be capable of both \emph{fusing} and \emph{emulating} modal features. By transforming the input of one modality into the features of another through simple and efficient operations, we can assume that the encoder can comprehend and align two distinct modalities well enough.

Based on the above analysis, in this paper, we \XP{propose a modal emulation-based multi-modal crowd counting approach} that leverages a two-pass learning paradigm to perform efficient modal emulation, alignment, and fusion, as illustrated in Figure~\ref{fig:promptd}. We conduct extensive experiments on two widely used RGB-T and RGB-D multi-modal crowd-counting benchmarks. The results show that our proposed method outperforms previous methods and demonstrates the effectiveness of our method in leveraging multi-modal information for crowd counting. The technical contributions can be further summarized as follows:

\begin{itemize} 

\item We propose a two-pass learning paradigm for multi-modal crowd counting. In addition to the normal \emph{multi-modal inference} pass, we propose a \emph{cross-modal emulation} pass that encourages the model to coordinate different modalities. This two-pass paradigm makes our approach distinct from traditional methods.

\item We propose a modality alignment loss to align the outputs of the two passes and bridge the semantic gap between different modalities. 

\item We develop a hybrid cross-modal attention module, which consists of a straight attention mechanism that focuses more on global attention and a modulated attention mechanism that emphasizes local attention, to enhance multi-modal fusion power.

\end{itemize}

\section{Related Work}

\subsection{Multi-modal Crowd Counting}

Currently, the crowd counting task has been extensively studied~\cite{liu2018decidenet,idrees2018composition,zhang2019wide,liu2019crowd,ma2020learning,lin2022semi,lin2024gramformer,wang2022eccnas}. To enhance the counting accuracy, several works have introduced information from other modalities, such as thermal or depth~\cite{zhang2020uc,pang2020hierarchical,fan2020bbs,li2022rgb,wu2022multimodal,zhou2023mc,yi2023perspective,meng2024multi}. Lian et al.~\cite{lian2019density,lian2021locating} introduce a large-scale RGB-D crowd counting dataset and leverage a depth prior and a density map to improve the head/non-head classification in the detection network.  Zhang et al.~\cite{zhang2022spatio} adopt a CSCA method to effectively capture and integrate information from different modalities. Zhou et al.~\cite{zhou2022defnet} propose a dual-branch enhanced feature fusion network to fuse RGB-thermal features.  Liu et al.~\cite{liu2021cross} and Tang et al.~\cite{tang2022tafnet} introduce a three-stream network for multi-modal fusion. Zhou et al.~\cite{zhou2023mc} propose a multimodality cross-guided compensation coordination network to predict crowd density maps by complementing different modules. However, these multi-modal approaches do not fully explore the modality alignment issue.
 
\subsection {Transformer for Multi-modal}

Many Transformer-based methods were proposed for multi-modal tasks~\cite{yan2023cross,zhong2023anticipative}. \wch{Vilbert \cite{lu2019vilbert} and LXMERT \cite{tan2019lxmert} use the cross-attention mechanism to learn vision-and-language connections. zhu et al.~\cite{zhu2023multi} propose a multi-modal feature pyramid transformer that fuses different modalities by intra-modal and inter-modal feature pyramid transformer. Zhang et al.~\cite{zhang2023cmx} design a cross-modal feature rectification module to calibrate bi-modal features and a two-stage Feature Fusion Module to enhance the information interaction.}

\subsection{Prompting Learning}

Recently, prompting Learning has achieved great success in computer vision tasks \cite{wang2022learning,wang2022dualprompt,lee2023multimodal}. Zhu et al.~\cite{ViPT}  develop a visual prompt multi-modal Tracking framework for various downstream multi-modal tracking tasks by learning modal-relevant prompts. Li et al.~\cite{li2023efficient} utilize a prompting method to extract fusion representations between different modalities.

It is crucial to emphasize that none of the previously mentioned methods encompass the idea of cross-modal emulation, which constitutes the central focus of our paper. Consequently, the foundational motivation, the implementation, and the pertinent loss functions employed to fine-tune the prompts diverge markedly from those outlined in previous work.

\section{Methods}

\begin{figure*}[t]
  \centering
  \includegraphics[width=1\linewidth]{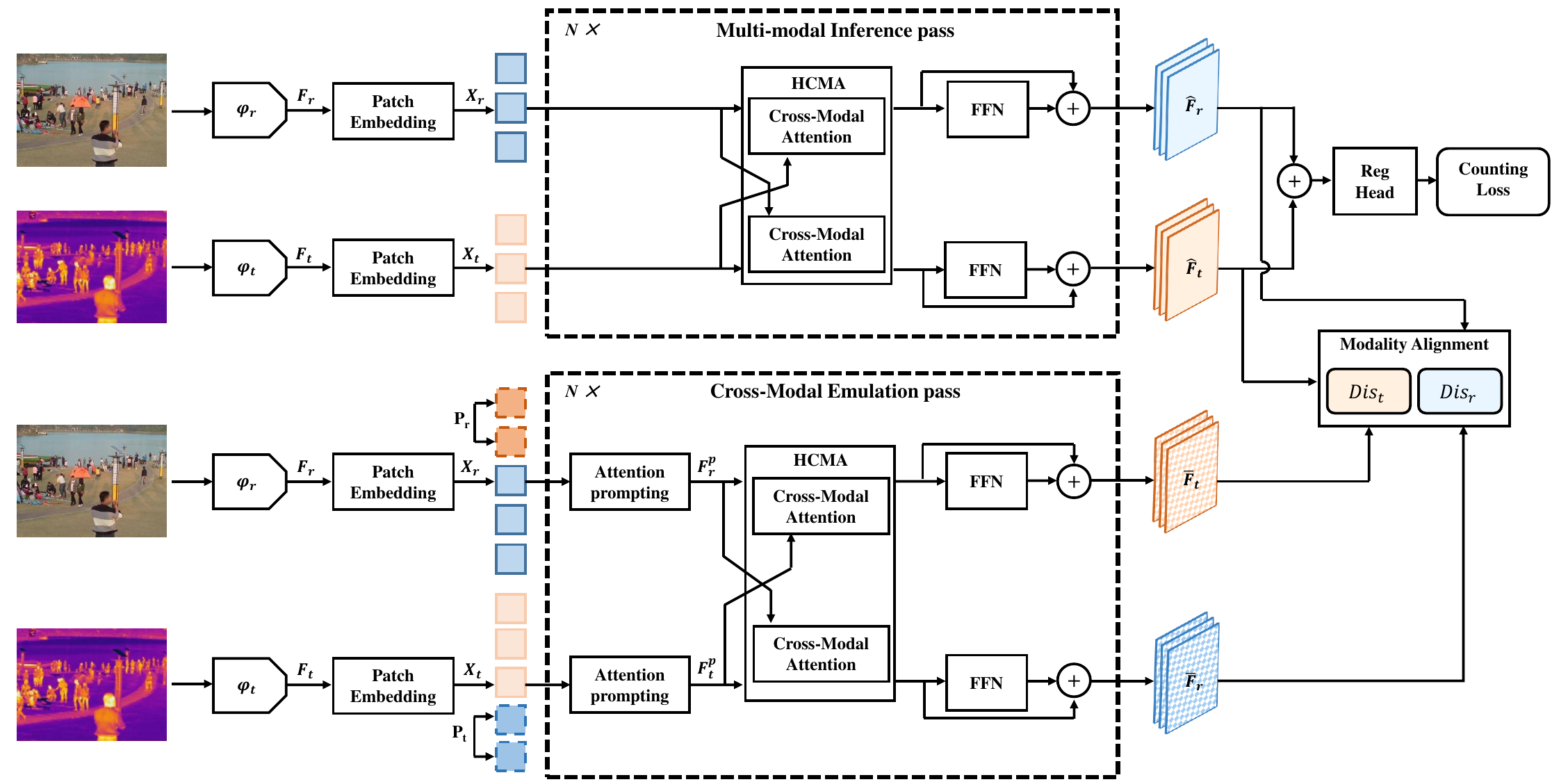}
\caption{\footnotesize \textbf{Illustration of the proposed framework}. Specifically, our framework consists of two passes: the Multi-modal Inference (MMI) pass and the Cross-modal Emulation (CME) pass. The MMI pass uses a hybrid cross-modal attention module to fuse global and local modalities. The CME pass shares the structure and weights with the MMI but emulates features of one modality into another, \emph{i.e.}, $F_r \rightarrow \bar{F_t}$ and $F_t\rightarrow \bar{F_r}$, using an additional attention prompting module. The process of emulation fosters the coordination of different modalities. Moreover, a loss function for modality alignment is employed to bridge the semantic gap that exists between these modalities.}
\label{fig:promptd}
\end{figure*}

Figure~\ref{fig:promptd} presents an overview of the framework, which mainly consists of a Multi-modal Inference (MMI) pass and a  Cross-modal Emulation (CME) pass. The two passes share most of the network structure and weights. And we place the proposed Hybrid Cross-modal Attention (HCMA) module behind each block of the dual-channel VGG19-like network ~\cite{liu2021cross}. 
Specifically, given an RGB image and a thermal image, to maintain the specific information of each modality, we feed them into the first three blocks of VGG19 \cite{simonyan2014very} $\varphi_{r}$ and $\varphi_{t}$ to extract modality-specific features of individual modality $F_{r}, F_{t} \in {R}^{C \times H \times W}$, where $C$, $W$, and $H$ are the channel, width, and height, respectively. \wch{And then, the $2D$ feature $F_{r}, F_{t}$ are embedded and flattened to a sequence of patch embeddings $X_{r}, X_{t}  \in {R}^{N \times D}$, where $N$ is the number of patches, and $D$ is the patch dimension.} To fully fuse the information of the two modalities, we introduce the MMI pass which consists of the HCMA module into the adjacent block of the VGG19 to capture global-local complementarity information. Meanwhile, the CME pass can modulate $F_{r}$ features into pseudo-thermal features $\bar{F_{t}}$ to enhance modality alignment. Similarly, the $F_{t}$ features can also be modulated into pseudo-RGB features $\bar{F_{r}}$. Next, the features produced by both passes are fed into the modality alignment loss, which aims to bridge the semantic gap across different modalities. Afterward, the output features of the MMI pass are linearly combined using a weighted sum and fed into a regression head to generate a prediction for the final high-fidelity crowd density map $\mathcal{\hat{D}}$. Finally, we combine the Bayesian Loss \cite{ma2019bayesian} to constrain the training of the overall model.

\subsection{Multi-modal \XP{Inference}} 

\begin{figure*}[t]

  \centering
 
  \includegraphics[width=0.9\linewidth]{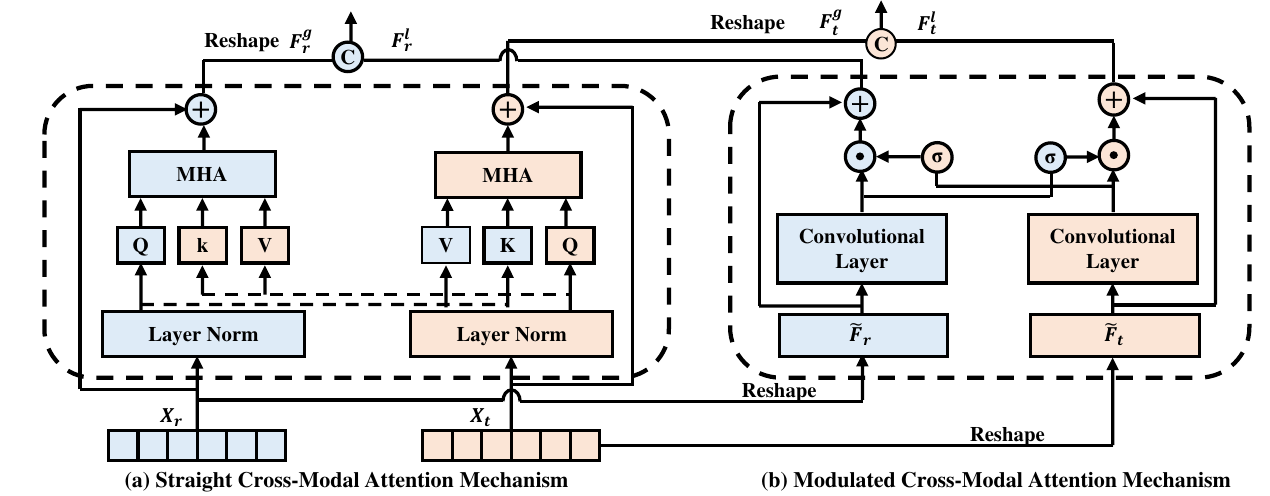}

\caption{\footnotesize Architecture of the Hybrid Cross-Modal Attention Module. (a) Straight Cross-modal Attention is used for global multi-modal fusion. (b) Modulated Cross-modal Attention is used to fuse local details, where the $\odot$, \textcircled{$\sigma$} and \textcircled{c} denote Hadamard product, Sigmoid function, and Concatenation operation, respectively. }
\label{fig:HCAM}
\vspace{-1em}
\end{figure*}

\XP{In the Multi-modal Inference pass, we design the HCMA module, which} comprises two types of attention mechanisms: straight cross-modal attention and modulated cross-modal attention, as shown in Figure~\ref{fig:HCAM}.

\noindent \textbf{Straight Cross-modal Attention}

To capture long-range contextual information by fusing global information from both modalities, we introduce the Straight Cross-modal Attention (SCMA) mechanism based on Multi-head Attention (MHA)~\cite{dosovitskiy2020image}, as shown in Figure~\ref{fig:HCAM} (a). \wch{Specifically, different modal patch embeddings $X_{r}$ and $X_{t}$ are linearly projected to produce their queries, keys, and values, respectively, which are denoted as $Q_{r}, K_{r}, V_{r}$ and  $Q_{t}, K_{t}, V_{t}$.} \wch{Then, we perform straight cross-modal attention, which can be calculated as follows}:

\begin{equation}
     H_{r} = s(\frac{Q_{r} K_{t}^T}{\sqrt{d}})V_{t}, \   H_{t} = s(\frac{Q_{t} K_{r}^T}{\sqrt{d}})V_{r},
\end{equation}

\noindent where $H$ terms the output of the attention head, $s$ terms the softmax function, and $\frac{1}{\sqrt{d}}$ is the scaling factor based on the query/key dimension $d$.  Finally, the outputs of each head are concatenated and fed to a series of operations including dropouts and residual concatenation, and then reshaped into 2D features to obtain the fused global features $F_{r}^{g}$ and $F_{t}^{g}$, where $F_{r}^{g}, F_{t}^{g} \in {R}^{C^{'} \times H \times W}$.

\noindent \textbf{Modulated Cross-modal Attention} 

We introduce the Modulated Cross-modal Attention (MCMA) mechanism to fuse local details in different modalities and obtain modulated complementary features, as shown in  Figure~\ref{fig:HCAM} (b). First, the patch embeddings $X_{r}$ and $X_{t}$ are reshaped into 2D feature maps $\tilde{F_{r}}$ and $\tilde{F_{t}}\in {R}^{C^{'} \times H \times W}$, respectively. These feature maps are then fed to the modulated cross-modal attention mechanism to capture local complementary information:

\begin{equation}
        F_{r}^{l} = \phi_r(\tilde{F_{r}}) \odot \sigma(\phi_t(\tilde{F_{t}})) + \tilde{F_{r}},
        F_{t}^{l} = \phi_t(\tilde{F_{t}}) \odot \sigma(\phi_r(\tilde{F_{r}})) + \tilde{F_{t}},
\end{equation}

\noindent where $\phi$, $\odot$, and $\sigma$ denote a two-layer convolutional layer, Hadamard product, and Sigmoid function, respectively. 

To combine complementary global and local features, we simply concatenate them together along the channel dimension and introduce a convolutional layer with $1\times1$ convolution kernel to reduce the concatenated features to the $C^{'}$ dimension. Then, we feed them into a two-layer feed-forward network ($f$) to obtain the fused feature:

\begin{equation}
    \hat{F_{r}} = f_{r}([F_{r}^{g},F_{r}^{l}]), \ 
    \hat{F_{t}} = f_{t}([F_{t}^{g}, F_{t}^{l}]),
\end{equation}

\noindent where the [,] defines the concatenation operation.

We further diversify the contributions of different modality features and assign weights to each modality based on their importance or relevance. The two modalities are then fused as a weighted sum and input to the regression head~\XP{$\gamma$}:

\begin{equation}
    \mathcal{\hat{D}} = \gamma(\alpha  \hat{F_{r}} + \beta  \hat{F_{t}}),
\end{equation}

\noindent where \(\alpha,\beta \in [0,1]\) are learnable parameters.

\subsection{Cross-modal Emulation}

\XP{As we argued in the introduction, modal emulation, by which the feature of one modality is converted into the one of another modality, is an important means of allowing models to fully comprehend and align different modalities. Motivated by this idea, we propose a cross-modal emulation pass to realize cross-modal modulation between the RGB and thermal features. }

\XP{We design the CME based on attention prompting \cite{wang2022dualprompt} which inserts prompts to the multi-head self-attention layer}. We split the prompts $P_{r}$ of RGB modal features into sub-prompts $P_{r}^{k}$, $P_{r}^{v}$ with the same sequence length, and prepend them to the key $K^{p}_{r}$ and value $V^{p}_{r}$ vectors while keeping query $Q^{p}_{r}$ vectors. $Q^{p}_{r}, K^{p}_{r}$ and $V^{p}_{r}$ vectors are generated by the RGB features $X_r$. Then, we can define the function of attention prompting as:

\begin{equation}
     F_{r}^{\textit{p}} = s(\frac{Q^{p}_{r} [P_{r}^{k} , K^{p}_{r}]^T}{\sqrt{d}})[P_{r}^{v},V^{p}_{r}],
\end{equation}

Similarly, we can also get the attention prompting of thermal features:

\begin{equation}
    F_{t}^{\textit{p}} = s(\frac{Q^{p}_{t} [P_{t}^{k},K^{p}_{t}]^T}{\sqrt{d}})[P_{t}^{v},V^{p}_{t}],
\end{equation}

Finally, \XP{through the CME pass $\psi$, the RGB features can be transformed to resemble the thermal features,  which are called the \emph{pseudo} thermal features $\bar{F_{t}}$.} Similarly, we can also obtain \emph{pseudo} RGB features $\bar{F_{r}}$:

\begin{equation}
   \wch{[\bar{F_{t}}, \bar{F_{r}}] = \psi(F_{r}^{\textit{p}},F_{t}^{\textit{p}})}.  \ 
\end{equation}

By converting one modality to another, the CME pass effectively modulates the feature representations of different modalities and enhance their alignment to better fuse information from different modalities. Notably, the CME pass is executed only in the training phase. Therefore, it does not increase the model size and extra overhead in the testing phase.

\subsection{Overall loss function}

\textbf{Modality Alignment Loss.} To align the outputs of the two passes to bridge the semantic gap between modalities, we use a Consistency Loss:

\begin{equation}
    \mathcal{L}_{CL} = \sum_{i=1}^{N}(\textit{Dis}\left ( \hat{F^i_{r}}, \bar{F^i_{r}} \right  ) +  \textit{Dis}\left ( \hat{F^i_{t}}, \bar{F^i_{t}} \right)),
\end{equation}

\noindent where $N$ is the number of training samples, the $\textit{Dis}(\cdot)$ is the distance metric and we simply use the Euclidean distance in our experiments.

\textbf{Counting Loss.} We adopt the Bayesian Loss (BL) \cite{ma2019bayesian} for crowd counting:
\begin{equation}
    \mathcal{L}_{BL} = \sum_{i=1}^N |1-  \langle\mathcal{\hat{D}},\frac{\mathcal{N}(\mathcal{D}_i,\sigma ^2\mathbf{I}_{2 \times 2})}{\sum_{j = 1}^{N}\mathcal{N}(\mathcal{D}_j,\sigma ^2\mathbf{I}_{2 \times 2})}\rangle|,
\end{equation}
\noindent \(\mathcal{N}(\cdot, \cdot)\) is a Normal distribution centered at the \(i\)th head point \(\mathcal{D}_i\) with standard deviation \(\sigma \).

Finally, the overall loss is

\begin{equation}
    \mathcal{L} = \mathcal{L}_{BL}  +  \mathcal{L}_{CL}.
\end{equation}

\section{Experiments}

We conduct experiments on two challenging datasets. \textbf{RGBT-CC} contains $2,030$ RGB-T image pairs, each with the size of $640\times480$. We follow \cite{liu2021cross} and use $1,030$, $200$, and $800$ pairs for training, validation, and testing, respectively. \textbf{ShanghaiTechRGBD} is a large-scale RGB-depth crowd counting dataset of $2,193$ images~\cite{lian2019density}. Each sample includes both an RGB image and a corresponding depth map. $1,193$ samples are assigned to the training set and the remaining ones for testing.

\textbf{Implementation details.} We implement our model on the Pytorch framework with an NVIDIA RTX $3090$ GPU. The CME and MMI passes share most of the network structure and weights. CME is implemented by incorporating an attention prompting module, with five learnable prompts, before the first HCMA block of MMI.  In Figure~\ref{fig:promptd}, \XP{the number of HCMA blocks $N$ is set to 3.} \wch{ In our implementation, patch dimension $D$ is set to $768$.} The SCMA mechanism is set to $1$ layer with $4$ heads, the patch size of the first HCMA block is set to $2$, while the last two blocks were set to $1$. This model has $160$M parameters, which is considered as our \emph{base} model. \wch{We also design a \emph{small} model with $82$M parameters, where the patch dimension $D$ in the three HCAM blocks are set to $256$, $512$, and $512$, respectively.} In the training phase, we adopt Adam as the optimizer, the learning rate is set to $0.00001$.  We set the batch size to $32$ on the RGBT-CC dataset and batch size to $1$ on the RGB-D dataset, respectively. Normal data augmentation is applied to the input images, including random crop and flip. The input images are randomly cropped to $256\times256$ for RGBT-CC dataset and $1024\times1024$ for RGB-D dataset. The max training epoch is set to $1500$. The Root Mean Square Error (RMSE)~\cite{sindagi2019multi} and the Grid Average Mean Absolute Error (GAME)~\cite{guerrero2015extremely} are adopted to evaluate the performance.

\subsection{Comparison with State-of-the-Art Methods}
\begin{table*}[t]

\caption{\small Comparison with the state-of-the-art methods on RGBT-CC dataset.}
\centering

\footnotesize
\begin{tabular}{c|c|c|c|c|c|c}
\hline
Method                   & Venue & GAME(0)  & GAME(1)  & GAME(2)  & GAME(3)  & RMSE   \\ \hline
UCNet   \cite{zhang2020uc}           & CVPR2020       & 33.96            & 42.42            & 53.06            & 65.07            & 56.31          \\
HDFNet   \cite{pang2020hierarchical} & ECCV2020       & 22.36            & 27.79            & 33.68            & 42.48            & 33.93          \\
MVMS \cite{zhang2019wide}          & CVPR2019       & 19.97            & 25.10            & 31.02            & 38.91            & 33.97          \\
BBSNet   \cite{fan2020bbs}           & ECCV2020       & 19.56            & 25.07            & 31.25            & 39.24            & 32.48          \\
CmCaF   \cite{li2022rgb}             & TII2022        & 15.87            & 19.92            & 24.65            & 28.01            & 29.31          \\
IADM  \cite{liu2021cross}          & CVPR2021       & 15.61            & 19.95            & 24.69            & 32.89            & 28.18          \\
CSCA   \cite{zhang2022spatio}        & ACCV2022       & 14.32            & 18.91            & 23.81            & 32.47            & 26.01          \\
TAFNet   \cite{tang2022tafnet}       & ISCAS2022      & 12.38            & 16.98            & 21.86            & 30.19            & 22.45          \\
BL+MAT   \cite{wu2022multimodal}        & ICME2022       & 12.35            & 16.29            & 20.81            & 29.09            & 22.53          \\
DEFNet   \cite{zhou2022defnet}       & TITS2022       & 11.90            & 16.08            & 20.19            & 27.27            & 21.09          \\ 
MC$^3$Net   \cite{zhou2023mc} & TITS2023       & 11.47            & 15.06            & 19.40           & 27.95            & 20.59         \\ \hline
\textbf{Ours-small}                        &   & 11.68   & 16.12   & 20.58            & 28.42            & \textbf{19.06} \\ 
\textbf{Ours-base}                        &   & \textbf{11.23}   & \textbf{14.98}   & \textbf{18.91}            & \textbf{26.54}            & \textbf{19.85} \\ \hline
\end{tabular}

\label{tab:sota}
\end{table*}

\begin{table*}
\caption{\small Comparison with the state-of-the-art methods on ShanghaiTechRGBD dataset.}
\centering
\footnotesize

\begin{tabular}{c|c|c|c|c|c|c}
\hline
Method                             & Venue     		& GAME(0)  & GAME(1)  & GAME(2)  & GAME(3)  & RMSE   \\ \hline
UCNet \cite{zhang2020uc}            & CVPR2020       	& 10.81            & 15.24         & 22.04               & 32.98           & 15.70                   \\ 
DetNet \cite{liu2018decidenet}      & CVPR2018        	& 9.74             & -             & -                   & -               & 13.14                 \\ 
HDFNet \cite{pang2020hierarchical}  & ECCV2020         	& 8.32            & 13.93         &  17.97              & 22.62           & 13.01                   \\ 
CL \cite{idrees2018composition}     & ECCV2018       	& 7.32                 & -         & -                     & -              & 10.48                      \\ 
BBSNet \cite{fan2020bbs}            & ECCV2020        	& 6.26                 & 8.53         & 11.80                & 16.46       & 9.26                          \\ 
BL+MAT \cite{wu2022multimodal}      & ICME2022         	& 5.39                & 6.73         &  8.98                  & 13.66      & 7.77                           \\ 
RDNet \cite{lian2019density}        & CVPR2019       	& 4.96                 & -         & -                        & -           & 7.22                              \\ 
CSCA \cite{zhang2022spatio}         & ACCV2022        	& 4.39                 & 6.47         & 8.82                & 11.76        & 6.39                      \\  
IADM \cite{liu2021cross}           	& CVPR2021       	& 4.38                 &  5.95         & 8.02                & 11.02        & 7.06                         \\  
DPDNet \cite{lian2021locating}      & TPAMI2021       	& 4.23                 &   5.67         &  7.04                &  9.64       & 6.75                         \\  
PESSNet \cite{yi2023perspective}      & TITS2023       	& 4.10                 &   -         &  -                &  -       & 6.02                         \\ \hline
\textbf{Ours-small}                       &					& 4.73         & 6.48      & 9.74            & 16.44          & 6.88 \\ 
\textbf{Ours-base}                       &					& \textbf{3.80}         & \textbf{5.36}      & 7.71            & 12.57          & \textbf{5.52} \\ \hline
\end{tabular}

\label{tab:rgbd}
\end{table*}
 
\begin{figure*}

  \centering
  \includegraphics[width=0.85\linewidth]{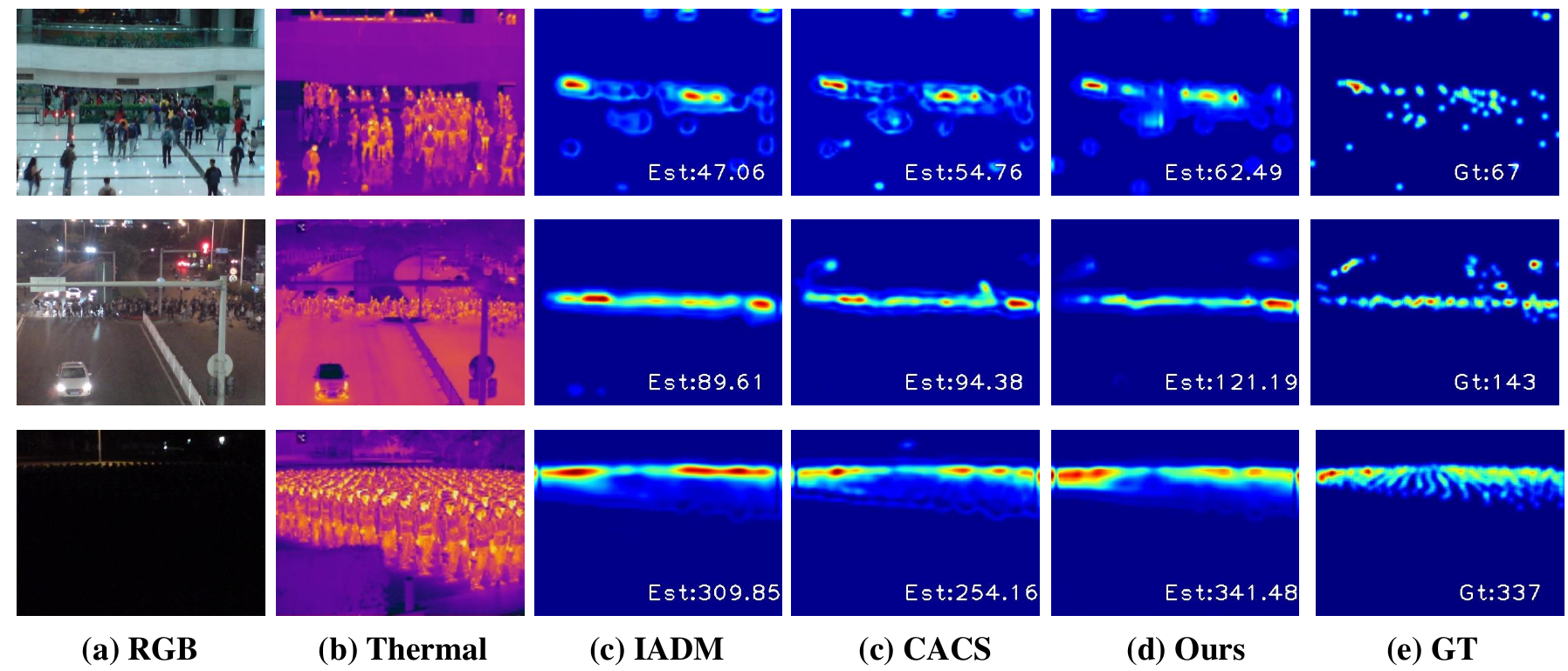}

\caption{\small Visualization results for generating crowd density maps with different models. }
\label{fig:SOTAmodal_visual}
\vspace{-3mm}
\end{figure*}

On the RGBT-CC dataset, the performance of all compared methods is shown in Table~\ref{tab:sota}. It could be found that the proposed method achieves better performance on evaluation metrics. For example, compared to MC$^3$Net, our model significantly improves counting performance, reducing GAME(0) to $11.23$ and RMSE  to $19.85$, respectively. Our model stands out due to its specific modal emulation ability to enhance the alignment and fusion of representations between RGB images and thermal images. We compare the visualization results of generating crowd density maps using different models to further validate our method in Figure~\ref{fig:SOTAmodal_visual}.

Moreover, we evaluate the ShanghaiTech-RGBD dataset. The results in Table~\ref{tab:rgbd} demonstrate that our method outperforms previous advanced models in terms of the main evaluation metrics (\emph{i.e.}, GAME(0), and RMSE). Specifically, our method achieves the lowest GAME(0) of $3.80$ and RMSE of $5.52$. The superior performance of our method demonstrates its generality and effectiveness in addressing multi-modal crowd counting tasks.

\subsection{Detailed Discussion}

\noindent \textbf{Ablation Study.}  We conducted a comprehensive study to evaluate the contribution of each component to the overall performance of the framework, as shown in Table~\ref{tab:GLCAP}. We start with \XP{the baseline, a two-stream expansion of the BL approach \cite{ma2019bayesian}}. \wch{Then, by incorporating the HCMA module consisting of SCMA and MCMA into the multi-modal inference process, our method consistently reduces the counting errors, specifically, by $6.25$ and $9.74$ in terms of GAME(0) and RMSE when compared with the baselines. The results demonstrate that the HCMA module contributes to enhancing the fusion of local information and global representation between the two modalities.} Furthermore, when using the CME pass with the attention prompting module, the best performance is achieved (\emph{i.e.}, GAME(0) is $11.23$, and RMSE is $19.85$). The success of the CME pass can be attributed to its ability to effectively coordinate information from different modalities. By aligning and fusing the modalities, the overall model performance is improved.

\begin{table*}[h] 

\caption{\small Ablation study on RGBT-CC dataset.}
\centering
\footnotesize

\begin{tabular}{cc|cc|c|c|c|c|c}
\hline
\multicolumn{2}{c|}{MMI}         & \multicolumn{2}{c|}{CME}     & \multirow{2}{*}{GAME(0) } & \multirow{2}{*}{GAME(1) } & \multirow{2}{*}{GAME(2) } & \multirow{2}{*}{GAME(3) } & \multirow{2}{*}{RMSE } \\ \cline{1-4}
\multicolumn{1}{c|}{SCMA} & MCMA & \multicolumn{1}{c|}{AP} & IP &                          &                          &                          &                          &                       \\ \hline
\multicolumn{1}{c|}{×}    & ×    & \multicolumn{1}{c|}{×}  & -  & 18.68                    & 22.91                    & 28.06                    & 35.87                    & 31.42                 \\
\multicolumn{1}{c|}{$\surd$}    & ×    & \multicolumn{1}{c|}{×}  & -  & 15.37                    & 20.63                    & 25.29                    & 33.01                    & 27.60                 \\
\multicolumn{1}{c|}{$\surd$}    & $\surd$    & \multicolumn{1}{c|}{×}  & -  & 12.43                    & 16.58                    & 21.26                    & 28.77                    & 21.68                 \\ \hline
\multicolumn{1}{c|}{$\surd$}    & $\surd$    & \multicolumn{1}{c|}{-}  & $\surd$  & 11.48                    & 15.95                    & 20.56                    & 28.57                    & 19.57                 \\
\multicolumn{1}{c|}{$\surd$}    & $\surd$    & \multicolumn{1}{c|}{$\surd$}  & -  & \textbf{11.23}                    & \textbf{14.98}                    & \textbf{18.91}                    & \textbf{26.54}                    & \textbf{19.85}                 \\ \hline
\end{tabular}

\label{tab:GLCAP}

\end{table*}

For the CME pass, we conduct experiments with different \XP{prompting techniques}, namely attention prompting (AP) and input prompting (IP)~\cite{lee2023multimodal}, in Table~\ref{tab:GLCAP}. Both \XP{prompting techniques work} for multi-modal crowd counting. However, we notice a trend where AP outperforms IP. This can be attributed to the fact that IP requires discarding the prompts in each HCMA block, which may potentially reduce its effectiveness.

We conduct experiments to compare HCMA with the vanilla cross-attention module in transformers as shown in Table~\ref{tab:VCAT}. \wch{Specifically, we replace the HCMA module with the cross-attention module and ensure that both models have a similar number of parameters (\emph{i.e.} $151$M for this model and $160$M for ours). The results indicate that the improvement achieved by our proposed method is not primarily attributable to the simple introduction of the cross-attention module or additional parameters, \XP{but instead benefited from} the carefully-designed HCMA and attention prompting modules.}

\begin{table}[h]

\caption{\small Comparison to the vanilla cross-attention (VCA) module.} 
\centering
\footnotesize
 
{
\footnotesize
\begin{tabular}{c|ccccc}
\hline
Model & GAME(0)  & GAME(1) & GAME(2)  & GAME(3)  & RMSE \\ \hline
VCA               & 15.11  & 19.66  & 24.29   & 31.92           & 27.94    \\ 
Ours              & \textbf{11.23}   & \textbf{14.98} & \textbf{18.91} & \textbf{26.54}       & \textbf{19.85}      \\  \hline
\end{tabular}}
\label{tab:VCAT}
\vspace{-0.5mm}
\end{table}

\begin{figure}

  \centering
  \centerline{\includegraphics[width=0.65\linewidth]{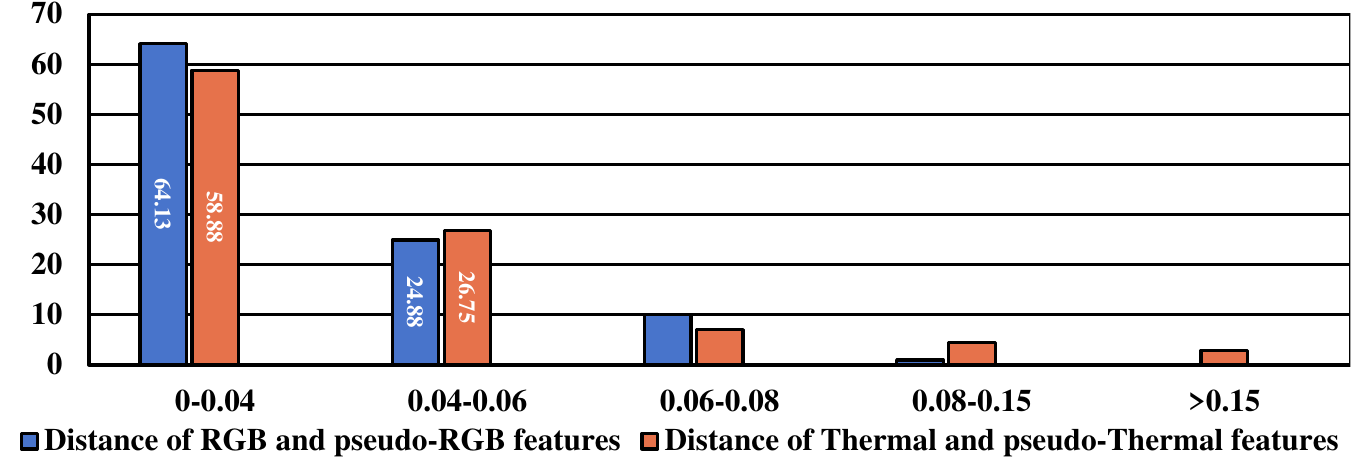}}

\caption{\small Distribution of the relative $L1$ distances between the real and pseudo features.}
\label{fig:MAE}
\vspace{-1em}
\end{figure}

\textbf{Effectiveness of CME pass.} \XP{We tabulate the distribution of the relative $L1$ distances between the thermal and pseudo-thermal features (as well as the RGB and pseudo-RGB features). The results are reported in Figure~\ref{fig:MAE}, where the horizontal axis indicates the ratio of the $L1$ distance to the average $L1$ norm of the real features, and the vertical axis represents the percentage of the test samples among $800$.} For thermal and pseudo-thermal features, $58.88$\% of the samples have the relative $L1$ distances below $0.04$, and $92.63$\% of the samples are below $0.08$. Similarly, for RGB and pseudo-RGB features, $64.13$\% of samples are below $0.04$, and $99.00$\% of samples are below $0.08$. This suggests that most of the pseudo samples only have a slight difference from the targeted samples. These results suggest that \XP{the CME pass well coordinates the two} modalities.

\begin{table}[h]
\caption{\small Impact of direct use of pseudo-features.}

\centering

{
\footnotesize
\begin{tabular}{c|ccccc}
\hline
         & GAME(0)  & GAME(1) & GAME(2) & GAME(3)  & RMSE  \\ \hline
W.   PF  & 12.21    & 16.39   & 21.12   & 28.70    & 19.98    \\ 
W/O  PF            & \textbf{11.23}   & \textbf{14.98} & \textbf{18.91} & \textbf{26.54}       & \textbf{19.85}    \\  \hline
\end{tabular}}
\label{tab:pseudo-features}
\vspace{-0.7mm}
\end{table}

Nevertheless, an additional inquiry arises: can these \emph{pseudo-modal features be directly employed} in generating the final density map output? We conduct an evaluation as shown in Table~\ref{tab:pseudo-features} to address it. When we concatenate the pseudo features and real features to feed the regression head, it can be found that the model performance is degraded. The rationale behind this phenomenon may lie in the fact that despite the pseudo-features closely approximating the features of the target modality, there still exist certain discrepancies. The direct amalgamation of one feature and its inferior counterpart does not lead to improvement but instead may result in a performance decline. \wch{In addition, if we use pseudo features for generating the density map, it means that the data needs to go through both the CME and MMI passes during testing. Although the MMI and CME passes share parameters, this will add extra overhead.}

\section{Conclusion}

We propose an effective emulation-based two-pass framework for multi-modal crowd counting. Our framework leverages a multi-modal inference pass that includes a hybrid cross-modal attention module, which fuses global and local complementary information from different modalities, as well as a cross-modal emulation pass that encourages the model to coordinate different modalities through attention prompting. Additionally, we introduce a modal alignment module to bridge the semantic gap between modalities. Through quantitative and qualitative experiments on RGB-T and RGB-D datasets, we demonstrate that our approach achieves competitive performance and high effectiveness for crowd counting. Our framework has promising potential to be applied to a variety of multi-modal tasks, which warrants further investigation in future research.

\section*{Acknowledgement}
This work was funded in part by the National Key R\&D Program of China (2021YFF0900500), the National Natural Science Foundation of China (62076195, 62376070, 62206271, 62441202, U22B2035), as well as the Fundamental Research Funds for the Central Universities (AUGA-5710011522).

\bibliography{main_bmvc_camera}
\end{document}